%% file: aaai2027.tex
\documentclass[letterpaper]{article} 
\usepackage[preprint]{aaai2027}
\usepackage[hyphens]{url}  
\usepackage{graphicx} 
\usepackage{natbib}  
\usepackage{caption} 
\usepackage{algorithm}
\usepackage{mdframed}
\usepackage{bm}
\usepackage{multicol}
\usepackage{multirow}
\usepackage{caption}
\usepackage{algorithmic}
\usepackage{amsthm}
\usepackage{booktabs} 
\usepackage{amsmath}
\usepackage{colortbl}
\usepackage{subfigure}
\usepackage{newfloat}
\usepackage{listings}
\usepackage{amsfonts}
\newtheorem{theorem}{Theorem}[section]
\newtheorem{proposition}[theorem]{Proposition}

\DeclareCaptionStyle{ruled}{labelfont=normalfont,labelsep=colon,strut=off} 
\floatstyle{ruled}
\newfloat{listing}{tb}{lst}{}
\floatname{listing}{Listing}

\usepackage{booktabs}

\title{	SciLT: Long-tailed Image Classification under Scientific Image Domains}
\author{
    Jiahao Chen and Bing Su
}
\affiliations{
    \textsuperscript{\rm 1}Renmin University of China
}

\begin{document}

\maketitle

\begin{abstract}
  Long-tailed recognition has benefited from foundation models and fine-tuning paradigms, yet existing studies and benchmarks are mainly confined to natural image domains, where pre-training and fine-tuning data share similar distributions. In contrast, scientific images exhibit distinct visual characteristics and supervision signals, raising questions about the effectiveness of fine-tuning foundation models in such settings. In this work, we investigate scientific long-tailed recognition under a purely visual and fine-tuning paradigm. Experiments on three scientific benchmarks show that fine-tuning foundation models yields limited gains, and reveal that penultimate-layer features play an important role, particularly for tail classes. Motivated by these findings, we propose SciLT, a framework that exploits multi-level representations through adaptive feature fusion and dual-supervision learning. By jointly leveraging penultimate- and final-layer features, SciLT achieves balanced performance across head and tail classes. Extensive experiments demonstrate that SciLT consistently outperforms existing methods, establishing a strong and practical baseline for scientific long-tailed recognition and providing valuable guidance for adapting foundation models to scientific data with substantial domain shifts.
\end{abstract}

\input{0.introduction}
\input{1.related_work}
\input{2.method}
\input{3.experiment}
\input{4.conclusion}

\bibliography{aaai2027}

\end{document}

%% file: 0.introduction.tex
\section{Introduction}
\begin{figure}
    \centering
    \includegraphics[width=0.99\linewidth]{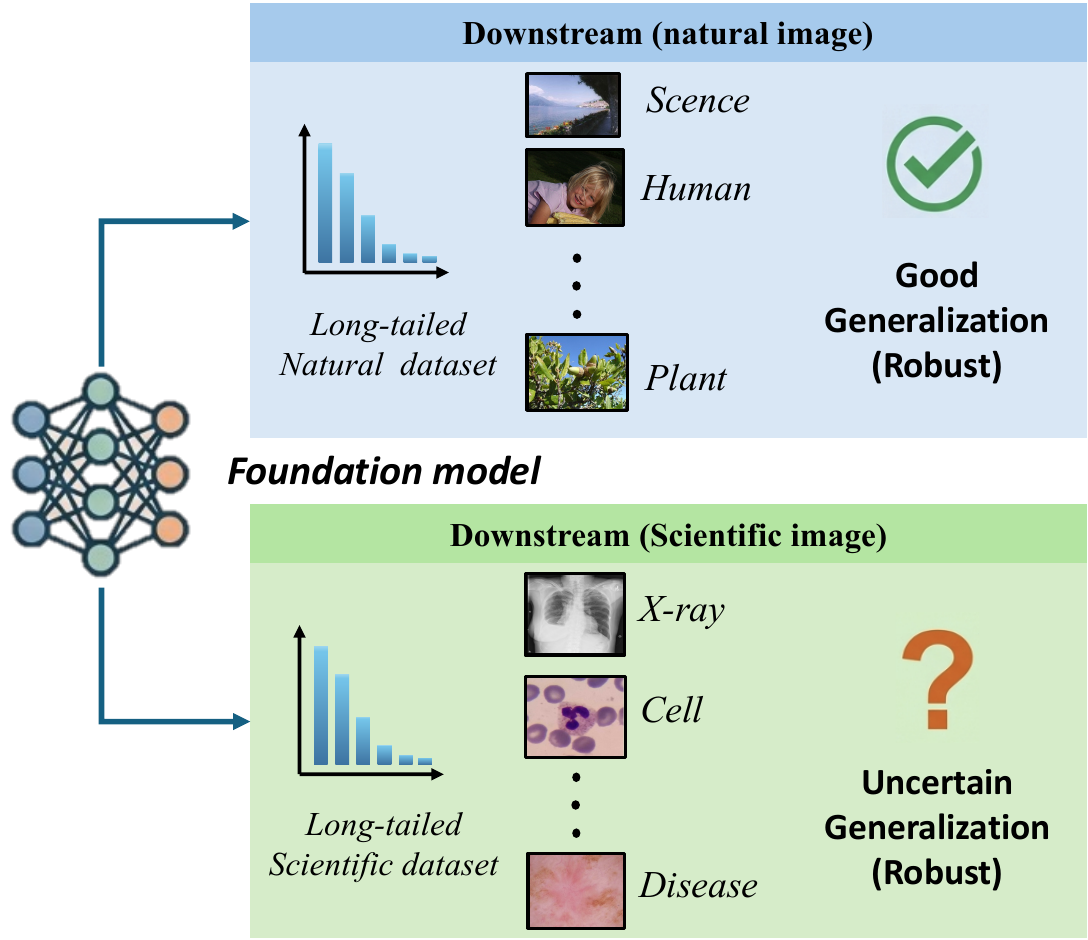}
    \caption{Differences in fine-tuning foundation models for downstream tasks on natural and scientific images. Fine-tuning achieves strong generalization performance on natural images (highlighted in blue), whereas its effectiveness on scientific images (highlighted in green) remains underexplored, due to the discrepancy in visual characteristics between scientific and natural images.}
    \label{motivation}
    \vskip -0.15in
\end{figure}

Real-world data often exhibit a long-tailed distribution, where most samples belong to a few head classes and only limited data are available for tail classes, leading to poor generalization and bias toward head classes~\cite{liu2019large,cui2019class} when training from scratch. Recently, the advent of foundation models (like ViT~\cite{dosovitskiy2020image}) has advanced long-tailed learning through pre-training followed by fine-tuning~\cite{shi2024long,dong2022lpt,tian2022vl}. However, existing benchmarks are confined to natural image datasets, where both pre-training and fine-tuning are conducted within the same domain. In contrast, many real-world scientific datasets also exhibit severe long-tailed distributions and are of critical importance, yet remain largely underexplored. As shown in Fig.~\ref{motivation}, scientific data differ substantially from natural images in two key aspects: (1) the visual characteristics and domain distributions exhibit pronounced discrepancies, leading to significant domain shifts; and (2) the downstream tasks and their required semantic information are fundamentally different, resulting in distinct supervision signals. These differences raise an open question as to \textit{whether fine-tuning foundation models remains effective under such scientific long-tailed scenarios}.

In this paper, we focus on a purely visual paradigm and systematically investigate parameter-efficient fine-tuning (PEFT), a typical adopted adaptation strategy for foundation models, for scientific long-tailed recognition, aiming to understand its behavior and develop effective solutions in domain-specific settings. As shown in Fig.~\ref{fig1}, we first observe that fine-tuning foundation models provides only limited benefits for tasks that deviate from the pre-training paradigm. Motivated by this observation, we conduct extensive experiments on three scientific long-tailed datasets, Blood, ISIC, and NIH-Chest, and further confirm that foundation models offer only marginal improvements in these scientific scenarios, as shown in Tab.~\ref{sci_performance} and Fig.~\ref{fig3}. Through a deeper investigation, we discover that features from penultimate layers can also contribute significantly to scientific long-tailed learning, and in certain cases, even outperform deeper layers, particularly for tail classes, as shown in Tab.~\ref{shallow} and Tab.~\ref{ab_m}. To better understand this phenomenon, we employ the Wasserstein distance to quantify the distributional discrepancy between penultimate and final-layer features, and identify a substantial gap between them, indicating that these layers capture markedly different information, as shown in Tab.~\ref{wass2}. Therefore, scientific long-tailed learning does not solely rely on the final-layer representations, and exploring features from other layers holds significance.

Based on this observation, we propose SciLT, a simple yet effective framework that exploits multi-level representation learning. Specifically, SciLT jointly leverages penultimate- and final-layer features through an adaptive fusion mechanism to construct more expressive representations, which are particularly beneficial for tail classes. During training, the fused features are optimized with Logit adjustment (LA)~\cite{menon2020long} criteria to explicitly address class imbalance, while the final-layer features are simultaneously trained with standard cross-entropy (CE) loss to preserve strong overall recognition capability. At inference time, predictions from the two complementary branches are ensembled, yielding a principled balance between head and tail class performance. After applying our method, we achieve a more balanced performance over all classes in different scientific long-tailed datasets. Our contributions can be summarized as follows:

(1) We systematically investigate the problem of scientific long-tailed recognition under the pre-training and fine-tuning paradigm. Through empirical analysis, we reveal that fine-tuning foundation models yields only limited benefits in scientific domains and uncover distinctive representation characteristics that differ substantially from those in natural image settings.

(2) We propose \textbf{SciLT}, a simple yet effective framework that exploits multi-level representations via adaptive feature fusion. By jointly optimizing penultimate- and final-layer features under different criteria, SciLT achieves a principled balance between head and tail class performance.

(3) Extensive experiments on three scientific long-tailed benchmarks, Blood, ISIC, and NIH-Chest, demonstrate that SciLT consistently outperforms existing methods, establishing a strong and practical baseline for scientific long-tailed recognition and providing valuable guidance for fine-tuning foundation models on scientific data with domain shifts.

%% file: 1.related_work.tex
\section{Related work}
\paragraph{Long-tailed learning.} 
Long-tailed learning methods address class imbalance by re-balancing training distributions via re-weighting~\cite{cui2019class} or re-sampling~\cite{ren2020balanced, guo2021long, kim2020imbalanced}, aiming to emphasize minority classes and mitigate bias. Representative approaches include class-balanced loss based on effective sample size~\cite{cui2019class} and logit adjustment using class priors~\cite{menon2020long}. More recently, large-scale pre-trained foundation models have demonstrated strong transferability and are increasingly applied to long-tailed recognition~\cite{dong2022lpt, shi2024long, tian2022vl, ma2021simple, zhao2025learning}. However, existing work mainly focuses on natural images and downstream distribution bias, with limited attention to the unique characteristics of scientific domains. In this paper, we study the fine-tuning of foundation models on imbalanced scientific datasets, systematically analyze the challenges under domain shift and long-tailed distributions, and explore targeted strategies for effective adaptation.

\paragraph{Scientific image representation learning}
Scientific data representation learning seeks transferable features from domain-specific data such as medical images, molecular structures, and materials micrographs. Early work adapts convolutional neural networks to medical imaging tasks~\cite{ragoza2017protein, jimenez2018kdeep}, while recent studies emphasize large-scale pre-training and self-supervised learning to improve generalization under limited supervision~\cite{raghu2019transfusion, azizi2021big}. More recently, class imbalance in scientific datasets has gained attention. Prior works address this via curriculum learning for multimodal diagnosis~\cite{han2025climd} and balanced contrastive frameworks for long-tailed recognition~\cite{zhaodeciphering}. However, these approaches remain largely task-specific, leaving the systematic adaptation of large-scale foundation models to imbalanced scientific domains underexplored. In this work, we conduct a comprehensive empirical study of foundation model transfer under severe domain shift and long-tailed distributions, revealing key limitations and providing principled guidance for effective adaptation.


%% file: 2.method.tex
\begin{table}
    \centering
    \small
    \tabcolsep=0.3cm    
    \begin{tabular}{l|ccc}
    \toprule
     & ImageNet-LT & Places365-LT & iNat2018 \\
    \midrule
    cRT     & 47.3 & 36.7 & 65.2 \\
    MiSLAS  & 52.7 & 40.4 & 71.6 \\
    PaCo    & 57.0 & 41.2 & 73.2 \\
    LiVT    & \textbf{60.9} & \textbf{41.2} & \textbf{76.1} \\
    \midrule
    \rowcolor{black!10}
    LPT     & -- & 50.1 \,(22\%) & 76.1 \,(0\%) \\
    \rowcolor{black!10}
    LIFT    & 77.0 \,(26\%) & 51.5 \,(25\%) & 79.1 \,(4\%) \\
    \rowcolor{black!10}
    VL-LTR  & 77.2 \,(27\%) & 50.1 \,(22\%) & 76.8 \,(1\%) \\
    \rowcolor{black!10}
    RAC & -- & 47.2 \,(15\%) & 80.2 \,(6\%) \\
    \bottomrule
    \end{tabular}
    \caption{Performance comparison on long-tailed benchmarks. Relative improvements are measured over the best of cRT, MiSLAS, PaCo, and LiVT. Gray rows denote fine-tuning from the foundation model.}
    \label{natural_performance}
    \vskip -0.1in
\end{table}

\begin{figure}
        \centering
    \subfigure[Places365-LT]{\includegraphics[width=0.45\linewidth]{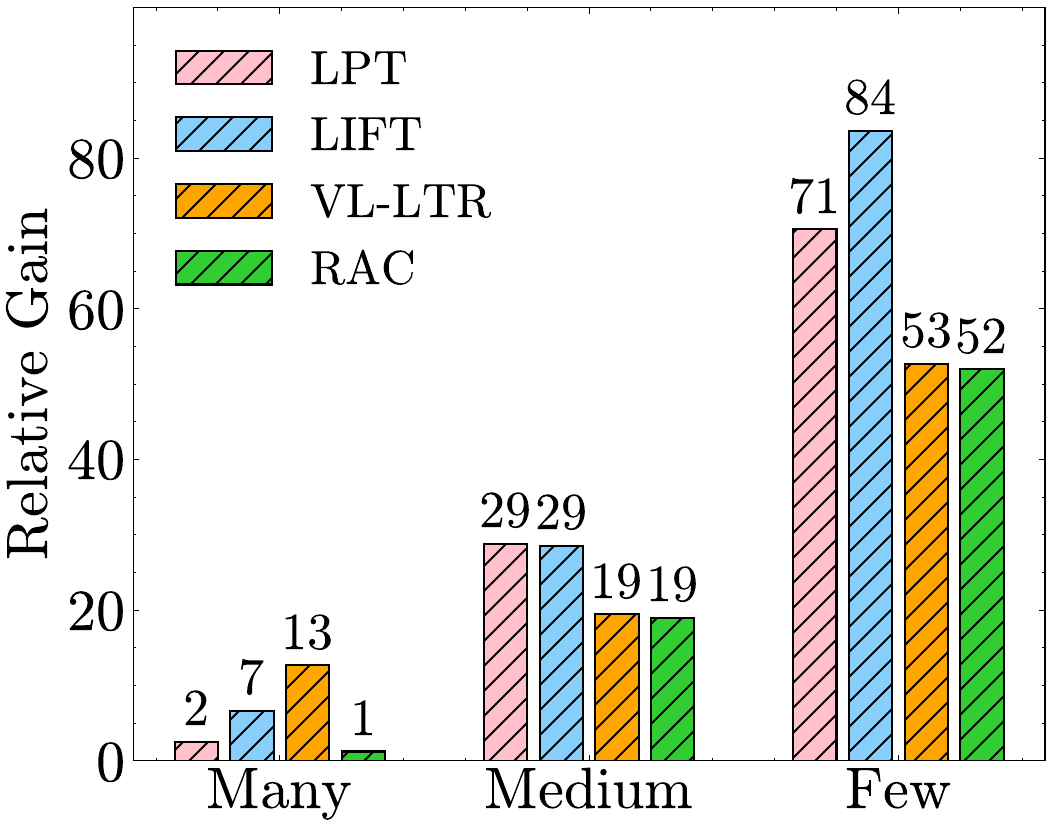}}
    \subfigure[iNaturalist2018]{\includegraphics[width=0.45\linewidth]{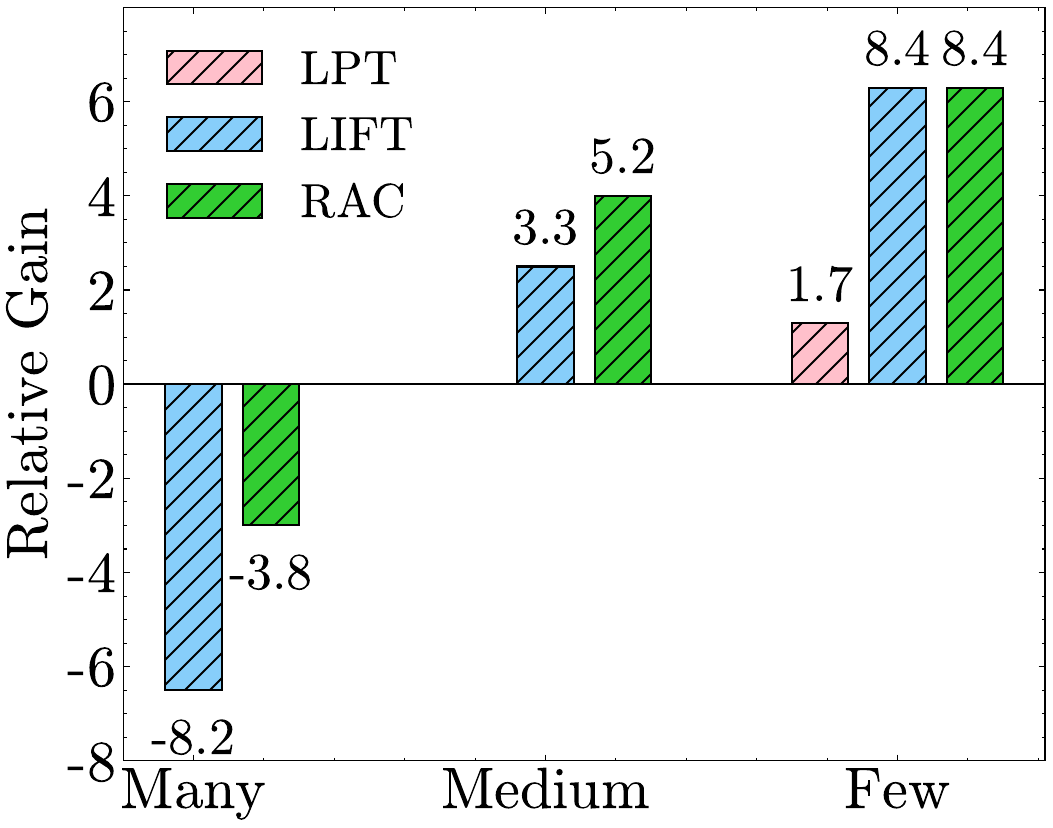}}
    \caption{Relative gain on (a) Places365-LT and (b) iNaturalist2018 datasets with ``Many", ``Medium", and ``Few" classes, respectively. RAC and LPT results on iNaturalist2018 are partially unavailable and therefore not fully plotted.}
    \label{fig1}
        \vskip -0.15in 
\end{figure}

\section{Motivation}
We first report the performance of representative long-tailed learning methods trained from scratch, including cRT~\cite{kang2019decoupling}, MiSLAS~\cite{zhong2021improving}, PaCo~\cite{cui2021parametric}, and LiVT~\cite{xu2023learning}, on widely used long-tailed benchmarks: ImageNet-LT, Places365-LT~\cite{liu2019large}, and iNaturalist2018~\cite{van2018inaturalist}. For comparison, we additionally evaluate approaches that fine-tune foundation models, such as LPT~\cite{dong2022lpt}, LIFT~\cite{shi2024long}, and VL-LTR~\cite{tian2022vl}. As shown in Tab.~\ref{natural_performance}, fine-tuning foundation models yields substantial performance gains on ImageNet-LT and Places365-LT, while only marginal improvements are observed on iNaturalist2018. Furthermore, Fig.~\ref{fig1} visualizes the relative performance gains on different fine-tuning methods across the \textit{Many}, \textit{Medium}, and \textit{Few} classes based on LiVT. The results indicate that fine-tuning foundation models provides more pronounced benefits for tail classes than for head classes on ImageNet-LT and Places365-LT. In contrast, iNaturalist2018 exhibits an atypical behavior, where the accuracy of head classes slightly degrades compared to training from scratch. We attribute this discrepancy to the intrinsic differences between these datasets: iNaturalist2018 is a fine-grained classification benchmark with over 8,000 classes, whereas Places365-LT and ImageNet-LT focus on more commonly encountered object and scene categories. Motivated by this observation, we further question whether the standard pre-training and fine-tuning paradigm remains effective for long-tailed tasks when there exists a large gap between the target domain and the pre-training domain, as well as a substantial mismatch in classification granularity and semantic structure.

\begin{mdframed}[backgroundcolor=gray!15] 
\noindent\textbf{Question 1:}  
\emph{How does the effectiveness of foundation model fine-tuning in long-tailed recognition depend on the domain and semantic granularity gap between pre-training and target datasets?}
\end{mdframed}


\section{Exploration on Scientific datasets}
Previous experiments indicate that fine-tuning foundation models does not yield sufficient performance gains on overall imbalanced datasets. In the following, we investigate the behavior of foundation model fine-tuning on long-tailed datasets, where the input domain and semantic granularity differ substantially from those of the pre-training data and the original optimization objectives.
\subsection{Settings}
\paragraph{Datasets} We conduct experiments on a diverse collection of scientific image datasets spanning multiple domains and task settings, including cellular images (Blood~\cite{tsutsui2023wbcatt}), radiographic images (NIH-Chest~\cite{wang2017chestx}), and disease-centric clinical images (ISIC~\cite{codella2019skin}). These datasets exhibit visual characteristics, imaging modalities, and semantic structures that are substantially different from those of natural image benchmarks commonly used for foundation model pre-training. In particular, they involve fine-grained medical semantics and domain-specific visual patterns, making them well suited for analyzing the limitations of foundation model fine-tuning under long-tailed and domain-shifted scenarios. We briefly summarize the core statistics of the datasets, with more detailed descriptions provided in Appendix Sec.B.

\paragraph{Experimental setup}
We conduct experiments using ViT-B/16~\cite{dosovitskiy2020image} pre-trained via CLIP~\cite{radford2021learning} on the Blood, ISIC, and NIH-Chest datasets, with each dataset randomly split into training, validation, and test sets. Following prior studies~\cite{shi2024long, dong2022lpt}, which indicate that parameter-efficient fine-tuning often outperforms full fine-tuning under long-tailed data distributions, we adopt AdaptFormer~\cite{chen2022adaptformer} to adapt the foundation model. To systematically analyze the impact of optimization objectives, we evaluate both the standard cross-entropy loss (CE) and the logit-adjusted loss (LA)~\cite{menon2020long} across all experimental settings. For evaluation, unlike ImageNet-LT or Places365-LT, which employ balanced test sets, we do not explicitly enforce test-set balance on scientific datasets, as their natural data distributions more closely reflect real-world scenarios. Accordingly, we report both overall accuracy (OvAcc) and Macro (mean-class) accuracy for each dataset. In addition, to isolate the effects of pre-training and fine-tuning, we train a ResNet~\cite{he2016deep} model from scratch (without any pre-training) under different loss configurations. This comparison enables a clearer assessment of the performance gains introduced by pretrained representations and parameter-efficient adaptation in imbalanced scientific image classification.

\subsection{Observation on scientific datasets}
\begin{table}
    \centering
    \small
    \tabcolsep=0.15cm
    \begin{tabular}{l|cc|cc|cc}
    \toprule
     & \multicolumn{2}{c|}{NIH-Chest}  & \multicolumn{2}{c|}{Blood} & \multicolumn{2}{c}{ISIC} \\
     \cmidrule{2-7}
    & OvAcc & Macro & OvAcc & Macro & OvAcc & Macro \\
     \midrule
     CE & \textbf{41.0} &13.6 &95.1 &94.7 &\textbf{75.3} &58.1\\
     CB &30.9 &17.3 &93.0 &95.0 &58.9 &55.1\\
     LDAM &41.0 &13.5 &\textbf{95.4} &\textbf{95.8} &73.8 &47.3\\
     LA & 22.2 &\textbf{24.6} &93.8 &\textbf{95.8} &66.5 &\textbf{65.9}\\
     Focal & 40.7 &13.4 & 94.3 &94.1 &74.5 &56.8\\
     LADE &23.4 &23.5 & 91.5 &92.3 &57.3 &48.1\\
    \midrule
    \rowcolor{black!10}
    CE &39.7 &11.1 & 98.2 &97.3 & 79.9& 62.2\\
    \rowcolor{black!10}
    LA &19.7 &20.8 & 97.6 &98.2 & 71.1& 72.4\\
    \bottomrule
    \end{tabular}
    \caption{Performance comparison on long-tailed scientific datasets. Rows shaded in light gray indicate results obtained via fine-tuning from the foundation model.}
    \label{sci_performance}
    \vskip -0.1in
\end{table}
\paragraph{Fine-tuning gains are dataset-dependent}
We evaluate the effectiveness of foundation models under both CE and LA objectives using AdaptFormer. For comparison, we also report the performance of training ResNet-18 from scratch with various long-tailed learning objectives, including CE, CB~\cite{cui2019class}, LDAM~\cite{cao2019learning}, LA, Focal~\cite{lin2017focal}, and LADE~\cite{hong2021disentangling}. As summarized in Tab.~\ref{sci_performance}, fine-tuning foundation models yields only marginal improvements on scientific datasets. Specifically, on ISIC, the overall accuracy increases from 75.3\% to 79.9\% (+4.6\%, \textbf{+6.4\% relative}), while on Blood, it improves from 95.4\% to 98.2\% (+2.8\%, \textbf{+2.9\% relative}). In contrast, fine-tuning on NIH-Chest fails to deliver consistent benefits and even underperforms training from scratch in terms of both overall and macro-averaged accuracy. In the following, we analyze the class-wise accuracy distributions.

\begin{figure}
    \centering
    \subfigure[NIH-Chest]{\includegraphics[width=0.49\linewidth]{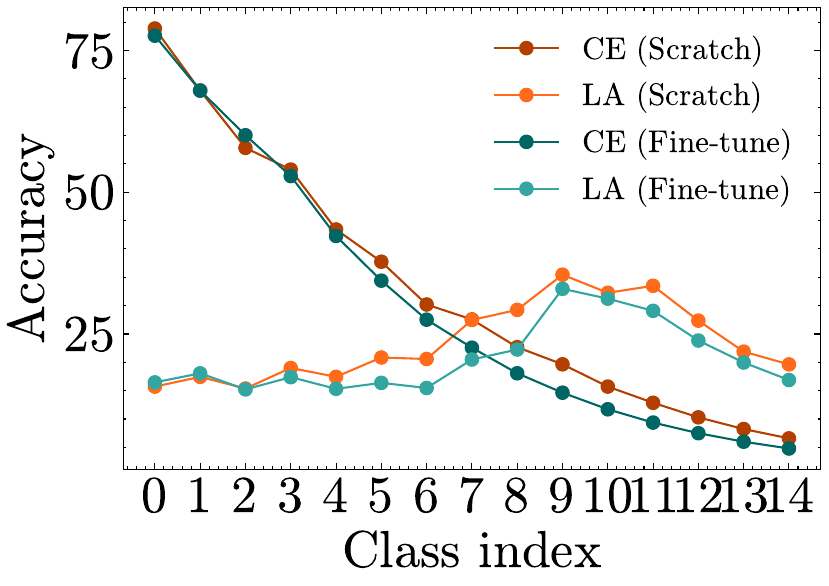}}
    \subfigure[ISIC]{\includegraphics[width=0.49\linewidth]{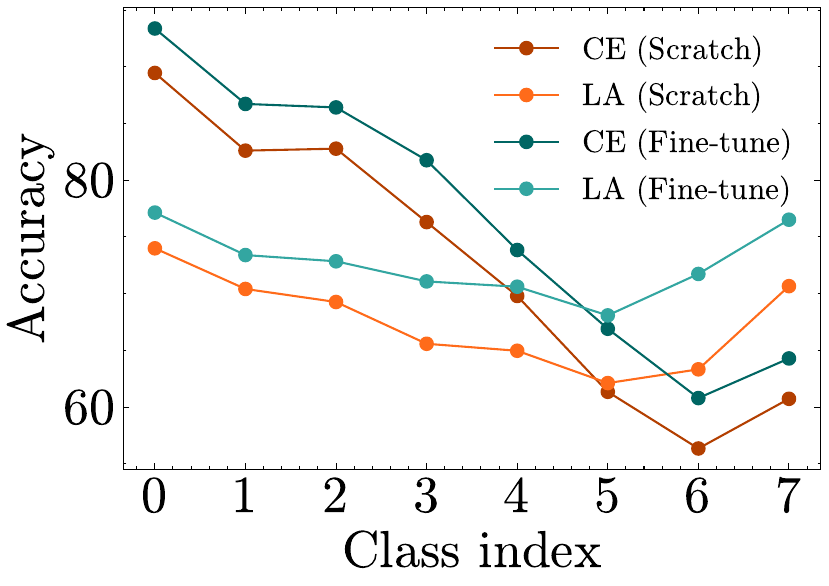}}
    \vskip -0.1in
    \caption{The performance curve on (a) NIH-Chest and (b) ISIC datasets with CE and LA training from scratch and fine-tuning. The class indices are sorted based on the number of samples belonging to each class. Curves are smoothed for better visualization.}
    \label{fig3}
    \vskip -0.1in
\end{figure}
As shown in Fig.~\ref{fig3}, on the NIH-Chest dataset, models trained from scratch consistently outperform fine-tuned counterparts across most categories under both CE and LA settings, indicating limited transferability of pretrained representations in this domain. Conversely, on ISIC, fine-tuning exhibits clear advantages, particularly for medium- and tail-class samples. Moreover, Fig.~\ref{fig3} also demonstrates that re-balancing objectives (e.g., LA) improves performance on tail classes, highlighting their critical role in mitigating severe class imbalance.

\begin{mdframed}[backgroundcolor=gray!15] 
\noindent\textbf{Finding 1:}  
\emph{Fine-tuning foundation models shows strongly dataset-dependent behavior on scientific long-tailed datasets, yielding non-trivial gains in some cases (e.g., Blood, ISIC) but potentially degrading under severe domain shift (e.g., NIH-Chest), while re-balancing objectives provide more consistent improvements for tail-class recognition.}
\end{mdframed}

\begin{table}
    \centering
    \small
    \tabcolsep=0.15cm
    \begin{tabular}{l|cc|cc|cc}
    \toprule
     & \multicolumn{2}{c|}{NIH-Chest}  & \multicolumn{2}{c|}{Blood} & \multicolumn{2}{c}{ISIC} \\
     \cmidrule{1-7}
    & OvAcc & Macro & OvAcc & Macro & OvAcc & Macro \\
     \cmidrule{1-7}
    CE $\dag$ &40.3 &11.9 & 97.9 &97.2 & 77.7 & 55.2\\
    LA $\dag$  &19.9 &20.2 & 96.8 &97.7 &  70.3 &68.8\\
    \bottomrule
    \end{tabular}
    \caption{Performance comparison on different long-tailed scientific datasets. $\dag$ denotes the performance of the penultimate layer.}
    \label{shallow}
    \vskip -0.1in
\end{table}
\begin{table}[t]
  \small
  \centering
  \begin{tabular}{l|l|ccc}
    \toprule
    & & Many & Medium & Few\\
    \cmidrule{2-5}
    \multirow{4}{*}{NIH-Chest} &CE &31.89&	1.33&	0.00\\
    &LA &15.58&	35.63&	11.33\\
    \cmidrule{2-5}
    &CE$\dag$ & 32.78&	2.75&	0.06\\
    &LA$\dag$ &16.60&	32.39&	14.24\\
    \bottomrule
  \end{tabular}
  \caption{The results on NIH-Chest with ``Many", ``Medium", and ``Few" (split via the number of samples of each class).}
  \label{ab_m}
    \vskip -0.1in
\end{table}

\paragraph{Penultimate layer can also benefit} Previous experiments indicate that fine-tuning foundation models can merely improve performance on long-tailed scientific datasets. Motivated by recent works~\cite{wang2022dualprompt, yang2025resclip, lan2024clearclip}, which show that penultimate layer of foundation models preserve informative representations and can support various downstream tasks, we further explore whether features from penultimate layer are beneficial for scientific long-tailed datasets. For details, we adopt AdaptFormer to fine-tune the foundation model, while extracting representations from the penultimate layer and attaching an additional classifier for final prediction. As shown in Tab.~\ref{shallow}, features from the penultimate layer achieve comparable, and in some cases superior, performance compared to those from the final layer. Notably, on NIH-Chest, the penultimate layer consistently yields higher overall accuracy and macro-averaged accuracy. Furthermore, we analyze performance across the \emph{Many}, \emph{Medium}, and \emph{Few} groups on NIH-Chest, and observe that the penultimate layer representations provide more improvements for tail classes than for head classes, as shown in Tab.~\ref{ab_m}. This behavior may stem from the scarcity of training samples in tail classes, which makes them more sensitive to the quality of feature initialization. By contrast, head classes benefit from sufficient data and can gradually learn task-specific representations during fine-tuning, thus relying less on pretrained features. Therefore, under scientific domain shifts, intermediate-layer representations may offer potential advantages.

\begin{mdframed}[backgroundcolor=gray!15] 
\noindent\textbf{Finding 2:}  
\emph{Penultimate-layer representations of foundation models provide more effective features for tail classes in scientific long-tailed datasets, yielding improved long-tailed recognition performance under severe domain shift.}
\end{mdframed}

\subsection{Feature space analysis} 
Previous experiments show that scientific datasets exhibit different behaviors from natural image benchmarks: fine-tuning foundation models yields only limited performance gains, and penultimate-layer representations can also benefit the long-tailed scientific learning. In this section, we further investigate the underlying reasons from a feature representation perspective. For details, we calculate the Wasserstein distance between the features of the penultimate and last layers.\footnote{We employ the Sinkhorn~\cite{cuturi2013sinkhorn} algorithm to estimate the Wasserstein distance, which is computationally efficient for high-dimensional feature distributions.} 

\begin{table}
    \centering
    \small
    \tabcolsep=0.06cm
    \begin{tabular}{l|ccc}
    \toprule
     & NIH-Chest & Blood& ISIC \\
     \midrule
    CE (penultimate-last) &0.98 &1.02 &0.96\\
    LA (penultimate-last) &1.01 &1.12 &0.97 \\
    \bottomrule
    \end{tabular}
    \caption{Wasserstein distance between the feature distributions of the penultimate and last layers under different training criteria.}
    \vskip -0.1in
    \label{wass2}
    \vskip -0.1in
\end{table}
As shown in Tab.~\ref{wass2}, a distributional discrepancy is observed between the penultimate and last layer features under different training criteria, indicating that the two layers encode distinct representation structures rather than forming a simple linear transformation. This discrepancy indicates that part of the information preserved in the penultimate layer is not fully retained in the final layer, which may lead to a degradation of discriminative signals during the final projection. Combined with previous experiments, we find that penultimate-layer representations provide complementary features for long-tailed scientific datasets, and effectively leveraging them is crucial for enhancing representation learning in this domain. 

\begin{mdframed}[backgroundcolor=gray!15] 
\noindent\textbf{Empirical observation 1:}
\emph{The two layers have different feature distributions, and their classifiers exhibit different class-wise errors. This combination motivates, but does not by itself prove, a benefit from feature fusion.}
\end{mdframed}

\section{Method}
The empirical study identifies uncertainty in both representation depth and class prior. SciLT constructs a retention-constrained representation from the two deepest layers and trains predictors at observed- and balanced-prior operating points. The two components address representation selection and class-prior bias, respectively.

\subsection{SciLT}
Consider input $\bm{x}\in\mathcal{X}$, label $y\in[C]$, class count $n_y$, and empirical prior $\pi_y$. Let $\bm{z}_{\ell}=f_{\ell}(\bm{x})\in\mathbb{R}^{d}$ denote the adapted representation at depth $\ell$. SciLT fuses $\mathcal{S}=\{N-1,N\}$ into $\widetilde{\bm{z}}$, then combines a fused balanced-prior predictor with a final-layer observed-prior predictor.

An unconstrained gate can prematurely suppress one depth for scarce classes. SciLT instead computes
$\kappa_\ell(\bm{x})=1/2+\sigma(\bm{w}_\ell^\top\bm{z}_\ell)$ and normalizes the two coefficients:
\begin{equation}
\begin{aligned}
\alpha_\theta(\bm{x})
&=\frac{\kappa_N(\bm{x})}
{\kappa_N(\bm{x})+\kappa_{N-1}(\bm{x})},\\
\widetilde{\bm{z}}
&=\alpha_\theta(\bm{x})\bm{z}_{N}
+[1-\alpha_\theta(\bm{x})]\bm{z}_{N-1}.
\end{aligned}
\label{eq:retention_agg}
\end{equation}
Because $\kappa_\ell\in(1/2,3/2)$, both weights lie in $(1/4,3/4)$, retaining each depth while allowing adaptation. The implementation forms the corresponding unnormalized feature before a cosine classifier; scale invariance makes the two forms prediction-equivalent.

Depth fusion does not remove the head-class bias induced by $\pi_y$. SciLT therefore uses two predictors:
\begin{equation}
\bm{s}^{\mathrm{bal}}=g_{\mathrm{bal}}(\widetilde{\bm{z}}),
\qquad
\bm{s}^{\mathrm{obs}}=g_{\mathrm{obs}}(\bm{z}_N).
\label{eq:two_predictors}
\end{equation}
For adjustment strength $\tau_h\ge0$, define
\begin{equation}
\ell_{\tau_h}(\bm{s},y)
=-\log
\frac{\exp(s_y+\tau_h\log\pi_y)}
{\sum_{c=1}^{C}\exp(s_c+\tau_h\log\pi_c)}.
\label{eq:prior_loss}
\end{equation}
The balanced and observed-prior branches use $\tau_{\mathrm{bal}}>0$ and $\tau_{\mathrm{obs}}=0$, respectively. Training minimizes
\begin{equation}
\mathcal{L}
=\lambda\ell_{\tau_{\mathrm{bal}}}(\bm{s}^{\mathrm{bal}},y)
+(1-\lambda)\ell_{\tau_{\mathrm{obs}}}(\bm{s}^{\mathrm{obs}},y).
\label{eq:general_objective}
\end{equation}
At inference, SciLT predicts from
$\bm{s}^{(\beta)}=\beta\bm{s}^{\mathrm{obs}}+(1-\beta)\bm{s}^{\mathrm{bal}}$.
The branches target class balance and observed-prior accuracy, respectively. Implementation details appear in the supplementary material.

\subsection{Theoretical Analysis}
We first formalize why the same representation depth need not be optimal for every class. For classification loss $\ell_{\mathrm{cls}}$, define the class-conditional depth risk and oracle depth as
\begin{equation}
\begin{aligned}
\mathcal{R}_{y,\ell}(g)
&=\mathbb{E}[\ell_{\mathrm{cls}}(g(\bm{z}_{\ell}),y)\mid Y=y],\\
\ell_y^\star
&\in\arg\min_{\ell\in\mathcal{S}}\mathcal{R}_{y,\ell}(g_\ell).
\end{aligned}
\label{eq:depth_risk}
\end{equation}

\begin{proposition}[Depth-dependent class risk]
\label{prop:depth_risk}
Suppose $\|\bm{z}_{\ell}\|_2\le R_\ell$, the linear classifier norm is at most $B$, and $\ell_{\mathrm{cls}}$ is $L$-Lipschitz in its logits. Let $d_{\mathrm{eff},y,\ell}$ denote the effective rank of the class-conditional feature covariance. For all $y$ and $\ell\in\mathcal{S}$, with probability at least $1-\delta$,
\begin{equation}
\begin{aligned}
\mathcal{R}_{y,\ell}(\widehat g_\ell)-\mathcal{R}_{y}^{\star}
\le{}&
\Delta^{\mathrm{shift}}_{y,\ell}
+\Delta^{\mathrm{approx}}_{y,\ell}
+\Delta^{\mathrm{opt}}_{y,\ell}\\
&+c_0LBR_\ell
\sqrt{\frac{d_{\mathrm{eff},y,\ell}
+\log(2C|\mathcal{S}|/\delta)}{n_y}},
\end{aligned}
\label{eq:depth_bound}
\end{equation}
where $c_0$ is universal and the three $\Delta$ terms denote transfer, approximation, and optimization error.
\end{proposition}

A deeper representation may reduce approximation error while increasing transfer bias or complexity. The $n_y^{-1/2}$ term amplifies this competition for rare classes. Proposition~\ref{prop:depth_risk} therefore explains why a global depth can be suboptimal, but does not establish that combining depths is beneficial. To characterize the value of fusion, consider a fixed shared linear readout, for which feature fusion induces the same convex combination of depth-specific scores. Under squared one-vs-rest loss, let $e_{y,N}$ and $e_{y,N-1}$ be the errors, with
\begin{equation}
\begin{aligned}
v_{y,N}&=\mathbb{E}[e_{y,N}^2\mid Y=y],\\
v_{y,N-1}&=\mathbb{E}[e_{y,N-1}^2\mid Y=y],\\
c_y&=\mathbb{E}[e_{y,N}e_{y,N-1}\mid Y=y].
\end{aligned}
\end{equation}

\begin{theorem}[Strict improvement from complementary errors]
\label{thm:complementarity}
If $c_y<\min\{v_{y,N},v_{y,N-1}\}$, the unique risk-minimizing weight is
\begin{equation}
\alpha_y^\star
=\frac{v_{y,N-1}-c_y}
{v_{y,N}+v_{y,N-1}-2c_y}\in(0,1),
\label{eq:oracle_alpha}
\end{equation}
and its squared surrogate risk is strictly smaller than that of either individual depth.
\end{theorem}

Theorem~\ref{thm:complementarity} identifies error complementarity, rather than feature discrepancy, as the source of the available fusion gain. Its optimal weight is a population quantity, however, and is difficult to estimate for tail classes. We therefore analyze the effect of restricting the learned gate. Let $\mathcal{R}_y(\alpha)$ be $L_\alpha$-Lipschitz on $[0,1]$, let $\alpha_y^\star$ minimize population risk, and let $\widehat\alpha_{y,a}$ minimize empirical risk on $[a_{\min},1-a_{\min}]$.

\begin{theorem}[Retention-constrained oracle regret]
\label{thm:retention_regret}
If
$\sup_{\alpha}|\widehat{\mathcal{R}}_y(\alpha)-\mathcal{R}_y(\alpha)|\le\varepsilon_y$, then
\begin{equation}
\mathcal{R}_y(\widehat\alpha_{y,a})-\mathcal{R}_y(\alpha_y^\star)
\le
2\varepsilon_y
+L_\alpha\left|
\Pi_{[a_{\min},1-a_{\min}]}(\alpha_y^\star)-\alpha_y^\star
\right|.
\label{eq:retention_regret}
\end{equation}
For a gate class of complexity $\mathfrak{C}_g$, with probability at least $1-\delta$,
\begin{equation}
\varepsilon_y
=O\!\left(
\sqrt{\frac{\mathfrak{C}_g+\log(1/\delta)}{n_y}}
\right).
\end{equation}
\end{theorem}

The first term quantifies gate-estimation uncertainty, whereas the second is the bias from clipping the oracle. Retention therefore trades controlled bias for protection against finite-sample gate collapse. Having characterized the representation component, we finally connect it to the two prior branches. Assume each branch reaches a calibrated population optimum in its representation space:
\begin{equation}
s_y^{(h)}
=\log p(\bm{z}_h\mid y)+(1-\tau_h)\log\pi_y+c_h(\bm{z}_h),
\label{eq:branch_optima}
\end{equation}
where $\bm{z}_{\mathrm{obs}}=\bm{z}_N$, $\bm{z}_{\mathrm{bal}}=\widetilde{\bm{z}}$, and $c_h$ is independent of $y$.

\begin{proposition}[Representation-and-prior geometric mixing]
\label{prop:geometric_mixing}
Mixed-logit prediction is equivalent to using the unnormalized class score
\begin{equation}
p(\bm{z}_N\mid y)^\beta
p(\widetilde{\bm{z}}\mid y)^{1-\beta}
\pi_y^{\,1-\beta\tau_{\mathrm{obs}}-(1-\beta)\tau_{\mathrm{bal}}}.
\label{eq:geometric_mixing}
\end{equation}
\end{proposition}

The first two factors combine representation evidence, while the last interpolates the effective prior. For $(\tau_{\mathrm{obs}},\tau_{\mathrm{bal}},\beta)=(0,1,1/2)$, it becomes $\pi_y^{1/2}$. SciLT is not pure prior interpolation because the branches use different representations. For independently trained nonlinear heads, Thm.~\ref{thm:complementarity} is a surrogate diagnostic. Detailed proofs are in Appendix Sec.E.

%% file: 3.experiment.tex
\section{Experiment}
We conduct experiments on Blood, ISIC 2019, and NIH-Chest following the settings described above, with additional details in Appendix Sec.C. We compare our method with cRT~\cite{kang2019decoupling}, MiSLAS~\cite{zhong2021improving}, PaCo~\cite{cui2021parametric}, BCL~\cite{zhu2022balanced}, LPT~\cite{dong2022lpt}, and LIFT~\cite{shi2024long}.
Overall accuracy (OvAcc) and macro-averaged class accuracy (Macro) are the primary metrics. OvAcc measures performance under the observed test prior, whereas Macro gives each class equal weight and is therefore insensitive to test-set class frequencies. We additionally report BalancedScore (BScore),
\begin{equation}
\mathrm{BScore}=\frac{2\,\mathrm{OvAcc}\,\mathrm{Macro}}
{\mathrm{OvAcc}+\mathrm{Macro}},
\label{eq:bscore}
\end{equation}
as a secondary descriptive summary. BScore is the harmonic mean of the two primary metrics and becomes small when either component is small. It does not replace balanced accuracy, Macro, or the Many/Medium/Few breakdown, and we do not claim that it is invariant to the test prior.

\subsection{Results}

\begin{table}[t]
    \centering
    \small
    
    \tabcolsep=0.05cm
    \begin{tabular}{l|cccccccc|c}
    \toprule
         & MEL & NV & BCC & AK & BKL & DF & VASC & SCC &BScore\\
         \midrule

cRT&	62.3&	89.1&	80.4&	52.8&	64.2&	48.5&	86.9&	42.7&	71.2\\
MiSLAS&	63.8	&88.4	&79.6&	58.7	&65.3	&56.2	&89.5	&45.1	&72.8\\
PaCo&	65.1	&89.7&	81.2&	60.5&	66.0&	58.4&	91.6	&46.3&	73.6\\
BCL&	65.8	&88.9	&80.8&	61.8&	66.4	&57.6	&92.4	&47.0&	73.3\\
LPT&	66.4	&89.5	&81.0&	62.7&	66.1&	60.2&	93.6&	47.8&	74.0\\
      LIFT   & 58.4& 77.2& 70.7& 68.8& 64.0& 86.4& 95.7& 58.0 &71.7\\
      LIFT (CE)   & 60.1& 93.4 &85.3 &42.2 &63.1& 36.4 &78.3& 39.1 &69.9\\
      \midrule
      SciLT &67.8 & 86.5 &78.1& 64.1& 67.0 &63.6& 95.7 &44.9&\textbf{74.5}\\
      \bottomrule
    \end{tabular}
    \caption{Results on ISIC datasets. We present the class-wise accuracy and the BScore.}
    \label{isic-final}
    \vskip -0.1in
\end{table}
\paragraph{Results on ISIC}
As shown in Tab.~\ref{isic-final}, SciLT consistently outperforms both LA and CE across most categories, yielding the highest overall BScore of $\bm{74.5}$. In particular, SciLT achieves gains on several challenging minority classes. For MEL, SciLT improves accuracy by $+\bm{9.4}$ points over LA and $+\bm{7.7}$ over CE, reaching $\bm{67.8}$. On AK, SciLT outperforms CE by a large margin of $+\bm{21.9}$ points, achieving $\bm{64.1}$. Moreover, SciLT also delivers consistent gains on BKL and SCC. These improvements on underrepresented categories significantly enhance class-wise balance, leading to a higher BScore and demonstrating the effectiveness of SciLT in handling long-tailed distributions.

\begin{table}[t]
    \centering
    \small
    \tabcolsep=0.08cm
    \begin{tabular}{l|ccccc|c}
    \toprule
         & Baso. & Eosino. & Lympho. & Mono. & Neutro. &BScore\\
         \midrule
    cRT&	99.8&	99.5&	98.7	&91.0&	98.0&	97.5\\
MiSLAS	&100.0	&99.6&	98.8	&92.4	&97.7	&97.6\\
PaCo	&99.8	&99.5&	98.6	&92.8	&97.5&	97.4\\
BCL	&100.0	&99.7	&98.9	&93.1	&97.7	&97.7\\
LPT	&100.0	&99.6	&98.8	&93.4	&97.6	&97.6\\
      LIFT   & 100&    99.7&  98.4 & 96.2 & 97.0& \textbf{97.9}\\
      LIFT (CE)   &100 &   99.7&  99.1 & 89.3&  98.4& 97.7\\
      \midrule
      SciLT &100&    99.7 & 98.5 & 93.6 & 97.5& 97.8\\
      \bottomrule
    \end{tabular}
        \caption{Results on Blood datasets. We present the class-wise accuracy and the BScore. Baso., Eosino., Lympho., Mono., and Neutro. denote basophil, eosinophil, lymphocyte, monocyte, and neutrophil, respectively.}
    \label{blood-final}
    \vskip -0.1in
\end{table}
\paragraph{Results on Blood}
As shown in Tab.~\ref{blood-final}, SciLT exhibits strong and stable performance on the Blood dataset, achieving a competitive BScore of $\bm{97.8}$. While all methods perform well on dominant classes, SciLT demonstrates clear advantages on the minority Mono. class, improving accuracy by $+\bm{4.3}$ points over CE and achieving $\bm{93.6}$. Meanwhile, SciLT maintains consistently high accuracy on other categories, including Baso., Eosino., Lympho., and Neutro., indicating that the proposed method effectively enhances minority class recognition without sacrificing head-class performance. These results highlight the robustness and balanced learning capability of SciLT under moderately long-tailed distributions.

\begin{table}[t]
  \small
  \tabcolsep=0.08cm
  
  \centering
  \begin{tabular}{l|ccc|ccc}
    \toprule
    & Many & Medium & Few & OvAcc &Macro &BScore\\
    \midrule
    cRT&	29.6&	8.7&	2.1	&34.5	&14.1	&20.0\\
MiSLAS&	30.4&	12.1&	3.7&	34.0&	15.8&	21.6\\
PaCo&	31.6&	12.8&	4.0&	34.7&	16.2&	22.1\\
BCL	&31.2&	14.2&	4.7&	34.2&	17.3&	23.0\\
LPT	&32.4&	14.9&	5.3&	35.3&	17.7&	23.6\\
    LIFT &15.58&	35.63&	11.33 &19.7& 20.8  &20.2   \\
    LIFT(CE) &31.89&	1.33&	0.00 &39.7 &11.1 &17.3\\
    \midrule
    SciLT & 33.5 & 16.78 & 6.07 &36.3 &18.8 & \textbf{24.8}\\
    \bottomrule
  \end{tabular}
    \caption{Results on NIH-Chest with ``Many", ``Medium", and ``Few" (split via the number of samples of each class).}
    \label{nihfinal}
\end{table}
\paragraph{Results on NIH-Chest}
As shown in Tab.~\ref{nihfinal}, SciLT substantially improves long-tailed recognition on NIH-Chest, particularly for the Medium and Few groups. For clearer comparison, we group the disease classes into ``Many'', ``Medium'', and ``Few'' according to their training sample sizes. Compared with CE, SciLT improves Medium and Few accuracy by $+\bm{15.5}$ and $+\bm{6.1}$ points, respectively, while increasing Macro by $+\bm{7.7}$ points at a $3.4$-point cost in OvAcc. Compared with LA, SciLT improves OvAcc by $+\bm{16.6}$ points while remaining $2.0$ points lower in Macro. SciLT achieves the highest BScore of $\bm{24.8}$, exceeding CE and LA by $+\bm{7.5}$ and $+\bm{4.6}$ points, respectively. These results indicate that SciLT preserves strong head-class and overall performance while recovering medium- and tail-class accuracy, yielding a better OvAcc--Macro trade-off rather than uniform superiority over either baseline.


\subsection{Ablation study}
\begin{table}[t]
\small
    \centering
    \begin{tabular}{l|ccc}
    \toprule
    &OvAcc &Macro &BScore \\
    \midrule
    w/o Fusion     &34.9&  15.1 & 21.1 \\
    SciLT            &36.3 &18.8 & 24.8  \\
    \bottomrule
    \end{tabular}
    \caption{Comparison of SciLT with the no-fusion configuration on the derived multiclass NIH-Chest task.}
    \label{ab_fus}
    \vskip -0.15in
\end{table}
\paragraph{Ablation study of the fusion model}To verify the effectiveness of our fusion strategy, we remove the fusion module and conduct ablation experiments by directly using the penultimate-layer features for classification and by ensembling the outputs of the two classifiers. As shown in Tab.~\ref{ab_fus}, the proposed fusion strategy brings consistent performance gains across all metrics: the overall accuracy improves by $\bm{+1.4}$ points (from $34.9$ to $36.3$), the macro accuracy increases by $\bm{+3.7}$ points (from $15.1$ to $18.8$), and the BScore exhibits a substantial gain of $\bm{+3.7}$ points (from $21.1$ to $24.8$). These results indicate that simple feature usage or output-level ensemble cannot fully exploit the complementary information between different representations, validating the necessity of the proposed fusion strategy.

\paragraph{Ablation study on the selected depths}
\label{sec:depth_ablation}
We examine whether progressively incorporating earlier backbone representations improves the depth-fusion configuration. We vary the selected depth set $\mathcal{S}$ on NIH-Chest while keeping the backbone, branch objectives, optimization schedule, inference rule, and seed fixed. For $|\mathcal{S}|$ depths, the mean anchor assigns coefficient $1/|\mathcal{S}|$ to each representation, followed by a depth-specific residual gate. As shown in Table~\ref{tab:depth_ablation}, the final-layer-only setting obtains 19.7 OvAcc, 20.8 Macro, and 20.2 BScore. Including the penultimate representation increases OvAcc to 36.3 and BScore to 24.8, although Macro decreases to 18.8. Adding one or two earlier layers yields BScores of 24.3 and 24.7, respectively, and neither setting exceeds the two-depth configuration. Thus, $\{N-1,N\}$ provides the highest observed OvAcc and BScore among the evaluated depth sets while using only two representations.

\begin{table}[t]
\centering
\small

  \tabcolsep=0.08cm
\begin{tabular}{l|ccc}
\toprule
Selected depths $\mathcal{S}$ & OvAcc & Macro & BScore \\
\midrule
$\{N\}$ &19.7  &20.8  &20.2   \\
$\{N-1,N\}$ & 36.3 &18.8 &24.8  \\
$\{N-2,N-1,N\}$ &35.7 &18.4 &24.3 \\
$\{N-3,N-2,N-1,N\}$ &36.2 &18.7 &24.7  \\
\bottomrule
\end{tabular}
\caption{Ablation of the selected depth set on NIH-Chest. The two-depth row is the current SciLT configuration.}
\label{tab:depth_ablation}
\vskip -0.1in
\end{table}



%% file: 4.conclusion.tex
\section{Discussion}
Despite the strong empirical performance of SciLT, several limitations remain. The current design primarily exploits representations from the penultimate layer, while richer interactions across multiple layers may further improve representation learning. More importantly, SciLT is designed as a general fine-tuning framework for scientific images under long-tailed distributions, rather than being tailored to a specific domain such as medical imaging. This setting is also fundamentally different from conventional domain adaptation or domain generalization: we do not assume access to data from multiple source domains or unlabeled target-domain data. Instead, SciLT operates on labeled data from the target scientific domain, where both domain shift from foundation-model pretraining and long-tailed class imbalance are present. Accordingly, the evaluated datasets are used as representative scientific benchmarks exhibiting these challenges, rather than to restrict the method to particular domains.

We also do not compare with certain methods such as VL-LTR~\cite{tian2022vl} and RAC~\cite{rac}, as they rely on auxiliary textual information that may be difficult to obtain or reliably verify in scientific applications. Nevertheless, SciLT is intended as an exploratory study toward understanding how foundation models can be effectively fine-tuned under the joint challenges of domain shift and long-tailed distributions. Its additional computational overhead is marginal relative to the backbone, and the findings provide broader insights beyond the specific benchmarks.

\section{Conclusion}
We explore foundation-model fine-tuning for long-tailed medical images and introduce SciLT, which jointly addresses representation depth and class prior through retention-constrained fusion and prior-complementary supervision. Results on ISIC, NIH-Chest, and Blood demonstrate its effectiveness and support our analysis of depth uncertainty, complementary errors, and retention trade-offs. These findings suggest that effectively exploiting intermediate representations and class-prior information is important for adapting foundation models to imbalanced scientific data. Future work will further validate these mechanisms through multi-seed experiments and extend SciLT to richer multi-layer interactions and broader scientific domains.